\documentclass[10pt,twocolumn,letterpaper]{article}

\usepackage{cvpr}              % To produce the CAMERA-READY version
\usepackage{graphicx}
\usepackage{amsmath}
\usepackage{amssymb}
\usepackage{booktabs}

\usepackage{multirow}
\usepackage[pagebackref,breaklinks,colorlinks]{hyperref}

\usepackage{titling}
\date{}
\predate{}
\postdate{}
\usepackage[resetlabels]{multibib}
\newcites{Main}{References}%
\newcites{Supp}{References}%

\usepackage[capitalize]{cleveref}
\crefname{section}{Sec.}{Secs.}
\Crefname{section}{Section}{Sections}
\Crefname{table}{Table}{Tables}
\crefname{table}{Tab.}{Tabs.}

\renewcommand\footnotemark{}

\def\cvprPaperID{2434} % *** Enter the CVPR Paper ID here
\def\confName{CVPR}
\def\confYear{2023}

\begin{document}

%%%%%%%%% TITLE
\title{\vspace{-0.2em} \Large \textbf{System-status-aware Adaptive Network for Online Streaming Video Understanding}   \vspace{0.3em}}

\author{Lin Geng Foo\textsuperscript{1\dag}\thanks{ \dag~Equal contribution;~~\S~Currently at Meta;~~\ddag~Corresponding author}
~~~ Jia Gong\textsuperscript{1\dag}
~~~ Zhipeng Fan\textsuperscript{2\S}
~~~ Jun Liu\textsuperscript{1\ddag} \\
\textsuperscript{1}Singapore University of Technology and Design ~~~~ \textsuperscript{2}New York University\\
{\tt\small \{lingeng\_foo,jia\_gong\}@mymail.sutd.edu.sg, zf606@nyu.edu, jun\_liu@sutd.edu.sg } \\
}

\maketitle

%%%%%%%%% BODY TEXT
\begin{abstract}
Recent years have witnessed great progress in deep neural networks for real-time applications. However, most existing works do not explicitly consider the general case where the device's state and the available resources fluctuate over time, and none of them investigate or address the impact of varying computational resources for online video understanding tasks. This paper proposes a System-status-aware Adaptive Network (SAN) that considers the device's real-time state to provide high-quality predictions with low delay. Usage of our agent's policy improves efficiency and robustness to fluctuations of the system status. On two widely used video understanding tasks, SAN obtains state-of-the-art performance while constantly keeping processing delays low. Moreover, training such an agent on various types of hardware configurations is not easy as the labeled training data might not be available, or can be computationally prohibitive. To address this challenging problem, we propose a Meta Self-supervised Adaptation (MSA) method that adapts the agent's policy to new hardware configurations at test-time, allowing for easy deployment of the model onto other unseen hardware platforms.
\end{abstract}

\section{Introduction}
Online video understanding, where certain predictions are immediately made for each video frame by using information in the current frame and potentially past frames, is an important task right at the intersection of video-based research and practical vision applications (e.g., self-driving vehicles \cite{gujjar2019classifying}, security surveillance \cite{emad2021early}, streaming services \cite{seow2008designing}, and human-computer interactions \cite{koppula2015anticipating}).
In particular, in many of these real-world video-based applications, 
a fast and timely response is often crucial to ensure high usability and reduce potential security risk. 
Therefore, in many practical online applications,
it is essential to ensure that the model is working with \emph{low delay} while maintaining a good performance,
which can be challenging for many existing deep neural networks.

Recently, much effort has been made to reduce the delay of deep neural networks, including research into efficient network design \cite{tan2021efficientnetv2,zhang2018shufflenet,howard2017mobilenets}, input-aware dynamic networks \cite{meng2020ar,habibian2021skip,fan2020adaptive,fan2021motion}, and latency-constrained neural architectures \cite{cai2018proxylessnas,lee2021help,cai2019once}. 
However, all these works do not explicitly consider the dynamic conditions of the hardware platform, and assume stable computation resources are readily available.
In practical scenarios, the accessible computing resources of the host devices can be \emph{fluctuating and dynamic} due to the fact that multiple computationally expensive yet important threads are running concurrently. 
For example, in addition to performing vision-related tasks such as object detection, human activity recognition, and pose estimation, state-of-the-art robotic systems usually need to simultaneously perform additional tasks like simultaneous localization and mapping (SLAM) to successfully interact with humans and the environment. 
Those tasks are also often computationally heavy and could compete with vision tasks for computing resources. 
As a result, at times when the host device is busy with other processes, conducting inference for each model might require significantly more time than usual, leading to extremely long delays, which could cause safety issues and lagging responses in many real-world applications. 
Therefore, the study and development of models providing \emph{reliable yet timely} responses under \emph{various} hardware devices and \emph{fluctuating} computing resources is crucially important. Unfortunately, such studies are lacking in the field.

To achieve and maintain \textit{low delay} for online video understanding tasks under a \emph{dynamic computing resource} budget, we propose a novel \textbf{S}ystem-status-aware \textbf{A}daptive \textbf{N}etwork (\textbf{SAN}). Different from previous works, SAN \textit{explicitly considers the system status} of its host device to make \textit{on-the-fly adjustments to its computational complexity}, and is thus capable of processing video streams effectively and efficiently in a dynamic system environment. 
SAN comprises of two components:
a) a simple yet effective dynamic main module that offers reliable predictions under various network depths and input resolutions;
b) a lightweight agent that learns a dynamic system-status-aware policy used to control the execution of the main module, which facilitates adaptation to the fluctuating system load.
With the adaptivity of the main module and the control policy generated by the agent, our SAN can achieve good performance on the online video understanding task while maintaining a low delay under fluctuating system loads.

In various applications, we may need to deploy SAN onto different hardware platforms for online video understanding.
However, it is inconvenient to train SAN for each hardware platform, and it might also be difficult to find adequate storage to load the large labeled dataset on all platforms (e.g., mobile devices).
In light of these difficulties, we further propose a method for \textit{deployment-time self-supervised agent adaptation}, which we call \textbf{M}eta \textbf{S}elf-supervised \textbf{A}daptation (MSA). 
With MSA, we can conveniently train a SAN model on a set of local platforms, and perform a quick deployment-time agent adaptation on a target device, without the need for the original labeled training data.
Specifically, our proposed \textbf{MSA} introduces an auxiliary task of delay prediction together with a meta-learning procedure, that facilitates the adaptation to the target deployment device.

In summary, the main contributions of this paper are:
\begin{itemize}
    \item We are the ﬁrst to explicitly consider the fluctuating system status of the hardware device at inference time for online video understanding. To address this, we propose \textbf{SAN}, a novel system-status-aware network that adapts its behavior according to the video stream and the real-time status of the host system.    
    \item We further propose a novel Meta Self-supervised Adaptation method \textbf{MSA} that alleviates the training burden and allows our model to effectively adapt to new host devices with potentially unclear computation profiles at deployment time.
    \item We empirically demonstrate that our proposed method achieves promising performance on the challenging online action recognition and pose estimation tasks, where we achieve low delays under a rapidly fluctuating system load without jeopardizing the quality of the predictions.        
\end{itemize}

\section{Related Work}
\textbf{Online Video Understanding.}
Recently, motivated by the increasing real-world demand, a lot of works \cite{zolfaghari2018eco,wu2019liteeval,nie2019dynamic,fan2021motion,fan2020adaptive} have attempted to improve the accuracy and efficiency of models for online video understanding tasks, such as online action recognition \cite{zolfaghari2018eco,lin2019tsm,wu2019liteeval} and online pose estimation \cite{nie2019dynamic,fan2021motion,habibian2021skip}.
Several works design \textit{efficient networks} to improve models' efficiency for online video tasks, including the efficient 3D CNN design \cite{zolfaghari2018eco}, the Temporal Shift 2D CNN network \cite{lin2019tsm} and the Skip-Convolution \cite{habibian2021skip},
while other researchers introduce adaptivity into the networks, termed as \textit{stream-aware dynamic networks} \cite{nie2019dynamic,meng2020arnet,fan2020adaptive} to further reduce the network's computation complexity while maintaining a good performance. 
For example, LiteEval \cite{wu2019liteeval} dynamically chooses between a coarse model and a fine model for each frame and MAPN \cite{fan2021motion} dynamically activates its encoder to save computation resources. 
However, all these existing methods do not explicitly consider the dynamic conditions of the hardware platform, and thus face limitations in the face of fluctuating system conditions.
For example, in situations when the system is under high computational load, the model may still select to execute the branch with high complexity and therefore incur significantly longer delays, which is sub-optimal for time-sensitive applications. 
Compared to these methods, our proposed SAN is system-status-aware and thus more robust to computational fluctuations on the models' host devices.
To the best of our knowledge, such a system-status-aware adaptive network that can efficiently tackle online video understanding tasks has not been explored before.

\textbf{Efficient Network Designs.}  
There are several paradigms for efficient deep architectures, and we list a few here. Efficient deep models such as SqueezeNet \cite{iandola2016squeezenet} and MobileNet \cite{howard2017mobilenets} introduce novel computation operators using fewer parameters, 
while Neural Network Quantization \cite{CouBen15Binaryconnect,zhou2017incremental} methods effectively shrink the model size of the existing models by quantizing  model parameters to isolated values. Teacher-student models \cite{hinton2015distilling} are also effective in distilling knowledge from larger models into smaller ones.
Some other latency-constrained networks \cite{ xu2021vipnas, wang2020hat} further conduct neural architecture search (NAS) to search for an efficient architecture.
Different from all these approaches, our SAN explicitly accounts for the dynamic system status that is constantly changing on the device over time, allowing us to perform efficient online video understanding even with a fluctuating system status.

\textbf{Reinforcement Learning (RL).}
RL methods \cite{mnih2015human,mnih2016asynchronous,lillicrap2015continuous} train agents to navigate environments such that the reward is maximized.
To allow pre-trained RL agents to adapt to a new environment in fewer episodes, methods based on the few-shot setting \cite{finn2017maml,jamal2019task} have been developed.
Differently, we aim to allow our RL agent to adapt at deployment time in a \textit{self-supervised manner} 
(i.e., no reward signals can be provided due to the absence of labeled video data).
We introduce MSA, a novel method for self-supervised deployment-time adaptation. MSA employs an auxiliary task of delay prediction, and performs meta-learning to allow the RL agent to effectively adapt its control policy to the target hardware platform via fine-tuning on the auxiliary task.

\section{Method}
Given the live video stream $V = \{I_1, ...,I_t,...\}$ and a seen/unseen host device whose system status $sys_t$ varies over time, our goal is to build a model that dynamically adjusts the computational complexity of inference based on not only the streaming input, but also the system status to make high-quality predictions with a low delay at each step $t$. 
To this end, our proposed model: 1)~takes into account the available computational resources to make on-the-fly decisions to dynamically adjust the computation consumption of the model, therefore maintaining the delay at low values while keeping the prediction's accuracy; and
2)~effectively adapts itself to unseen devices in a self-supervised manner while still keeping its performance during deployment.

\subsection{SAN: System-status-aware Adaptive Network}
\label{Sec: SAN}
\begin{figure*}
\centering
\includegraphics[width=0.8\linewidth]{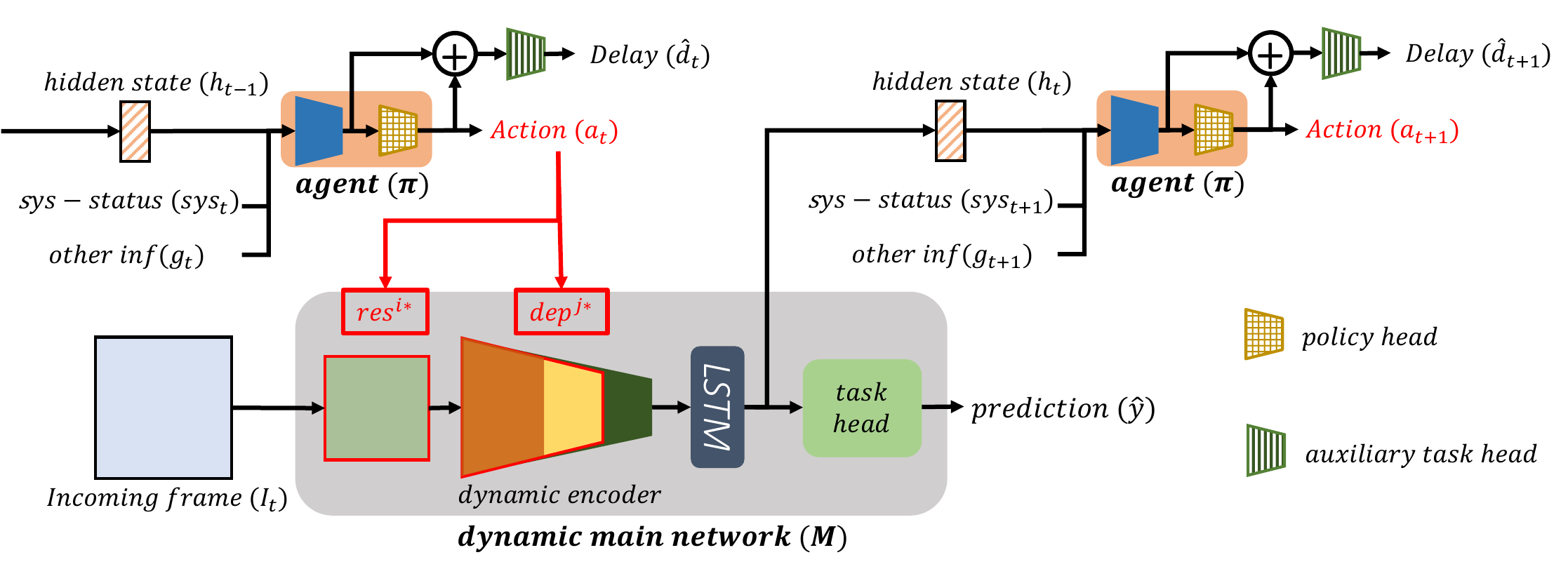}
\vspace{-4mm}
\caption{
Overview of our SAN framework, which consists of two modules: the agent module $\pi$ and the dynamic main network $M$. For each incoming frame $I_t$, the agent module first takes in the LSTM hidden state $h_{t-1}$, system-status $sys_t$, and other useful information $g_t$, and then decides the input resolution $res_t$ and the depth $dep_t$ of $M$ via the policy head. After that, the main network $M$ processes the frame according to these decisions. 
By controlling the behavior of the main network $M$ via the agent $\pi$, 
our SAN can output high quality and low delay predictions in the dynamic system environment. 
Note that the auxiliary task head of the agent is used for fine-tuning of the agent during deployment, calibrating the learned system-status-aware features to adapt to the unseen target device.
}
\vspace{-4mm}
\label{fig: pipeline}
\end{figure*}

For better online performance amid the host device's varying background computational load, 
we propose a System-status-aware Adaptive Network (SAN).
As shown in Fig.~\ref{fig: pipeline}, our SAN consists of two main components: a lightweight agent $\pi$ that decides how to process the incoming frame and a dynamic main network $M$ that processes the frame with the specified policy generated by $\pi$.

\textbf{Dynamic main network.}
Our dynamic main network $M$ flexibly produces frame-wise predictions using an architecture determined by the policy from the agent module $\pi$.
To accomplish this without extra burden on the hardware system, we design a \textit{dynamic encoder} that has both \textit{dynamic depth} and \textit{dynamic resolution}.
Our dynamic depth mechanism gives the encoder the option of producing output features at shallower intermediate layers without having to wait for all layers to be executed. This reduces computational cost by executing a ``simpler" sub-network.
On the other hand, the dynamic resolution mechanism allows our encoder to selectively take in input images at a lower resolution. 
Crucially, this reduces the amount of pixels to be examined by the network, effectively reducing the amount of convolution operations.
We implement both of the aforementioned mechanisms in a single network, where to enable dynamic execution depth of the network, we add early exits to the network which allow the execution to end at an intermediate layer, while to enable dynamic resolution, we leverage the fully convolutional network and add a dynamic global pooling operation at the end.

Specifically, as shown in Fig.~\ref{fig: pipeline}, our dynamic encoder selects from $m$ input resolutions $\{res^i\}_{i=1}^m$ and $n$ model depths $\{dep^j\}_{j=1}^n$, allowing for as many as $m \times n$ different options in total. These $m \times n$ dynamic options have different computation complexities and levels of performance.
For online video understanding, we construct our dynamic encoder as a 2D CNN for handling each frame, and to model the temporal aspect, the output features from the dynamic encoder are fed into a Long Short Term Memory network (LSTM) at every step.
The LSTM updates its hidden state $h_t$ based on the features of the incoming frame $I_t$ and its previous state $h_{t-1}$, which is then used by the task head to determine the final prediction $\hat{y_t}$ for the video task following: 
$ \hat{y_t} = M(I_t,res_{t},dep_{t},h_{t-1}; \theta)$,
where $\theta$ refers to the parameters of the main network $M$, while $res_{t}$ and $dep_{t}$ are selected by the agent introduced next. 
More details of our main dynamic network are in Sec.~\ref{sec: Experiments} and Supplementary.

\textbf{RL-based agent.}
The RL-based agent $\pi$
controls the dynamic main network $M$ by generating a frame-level processing policy $(res_{t}, dep_{t})$ at each time $t$, which aims to maintain the task-related accuracy $acc_t$ at high level with low delay $d_t$.
To make decisions amid the fluctuating system loads, our lightweight agent $\pi$ should be both \textit{system-status-aware} and \textit{streaming input-aware}.
We model our task as a Markov Decision Process, where the agent $\pi$ takes an action $a_t$ at every step $t$, based on the observed state $s_t$, and transferred to the next state $s_{t+1}$.
Crucially, state $s_t$ includes system status information $sys_t$ and input stream information $h_{t-1}$, making $\pi$ both system status and stream aware.
To learn an optimal policy, we propose an RL-based method to train our agent.

Specifically, we set the observed states of the agent as $s_t = [h_{t-1}, sys_t, g_t]$, where
$sys_t$ is the system status (e.g., the CPU and GPU utility)
\footnote{We also collect other system parameters (e.g., the temperature of each processor, the memory usage and the status of I/O)
via three linux utilities \cite{nvitop,psutil,pynvml} to form system status $sys_t$. More details are in Supplementary.}, 
$h_{t-1}$ is the LSTM hidden state, and $g_t$ refers to other useful information (i.e., the previous step's action $a_{t-1}$ and delay $d_{t-1}$).
Then, based on the available resolutions $\{res^i\}_{i=1}^m$ and depths $\{dep^j\}_{j=1}^n$, we build the action space as $\{a^{i,j} = (res^{i}, dep^{j})\}_{i=1,j=1}^{m,n}$. As shown in Fig. \ref{fig: pipeline}, at time $t$, the agent $\pi$ receives $h_{t-1}$, $sys_t$ and $g_t$, and generates a probability distribution over the action space:
\begin{equation}
P_t = [p^{1,1}_t, p^{1,2}_t,...,p^{m,n}_t] = \,\,\pi(h_{t-1}, sys_t, g_t;\phi)
\end{equation}
where $p^{i,j}_t \in [0,1]$ is the probability of the action $a^{i,j}$ at time $t$, $P_t\in [0,1]^{m \cdot n} $ represents the probability distribution over the action space, and $\phi$ refers to the parameters of the agent $\pi$.
Then, we take the highest probability entry in $P_t$ (denoted as $p^{i*,j*}_t$), 
select the corresponding action $a_t = (res^{i*},dep^{j*})$ and execute the main network accordingly.

\textbf{Training of SAN.}
To train our main network $M$, we apply the loss $\mathcal{L}_{M}$ based on the specified online video-processing task. 
For example, for online action recognition, $\mathcal{L}_{M}$ can be the cross-entropy loss, while for online pose estimation, $\mathcal{L}_{M}$ can be the Mean Squared Error (MSE) loss.

To control the dynamic encoder to achieve a good accuracy at low delay, we apply an RL-based loss $\mathcal{L}_{\pi}$ to train our agent $\pi$.
Specifically, we adopt a policy gradient loss \cite{williams1992simple,weaver2001optimal}.
To maximize the accuracy $acc_t$ while reducing the delay $d_t$, we set the reward of the agent at time $t$ as: 
\begin{equation}
\label{eq: agent_training}
r_{t} = \lambda_{acc} acc_t(a_{t},I_t)  - max \bigg( d_t(a_{t}, sys_t)-d_b, 0 \bigg)
\end{equation}
where $acc_t(a_{t},I_t)$ and $d_t(a_{t}, sys_t)$ are the accuracy and delay of our main network under the action $a_{t}$, the current frame $I_t$, and system status $sys_t$. $\lambda _{acc}$ is the coefﬁcient of the accuracy term, and $d_b$ is a tolerance threshold for the delay (e.g., $d_b$ = 30 $ms$ for a rate of 30 fps), which is the acceptable delay. The policy gradient loss $\mathcal{L}_{\pi}$ is defined as:
\begin{equation}
\mathcal{L}_{\pi} = \sum_{t=1}^T  - r_{t}  \log(p_t)
\end{equation}
where $p_t \in [0,1]$ is the probability of selecting the action $a_{t}$ at time $t$.

\subsection{MSA: Meta Self-supervised Adaptation}
\label{Sec: MSA}

Ideally, we want to deploy SAN onto many different hardware platforms for online video applications, requiring us to train SAN for each of those platforms, which is labor intensive, computationally prohibitive, and may also be infeasible at times due to storage concerns (e.g., mobile devices). 
Instead, a more practical option would be to effectively fine-tune a pre-trained SAN at deployment-time on a target device.
However, during deployment, the ground-truth labels associated with the real-time video stream are often inaccessible. To address this issue, we introduce an auxiliary task \cite{sun2020testtime,hansen2021selfsupervised} of \emph{delay prediction},
where the agent $\pi$ is expected to additionally predict the actual processing time based on the selected action (i.e., the input resolution and execution depth of main network $M$).

Our intuition is that, on a new (unseen) platform, our agent $\pi$ lacks understanding of hardware status of this platform, thus we need a way for the agent to quickly understand it.
Hence, we introduce the auxiliary task of delay prediction, because delay times are directly related to the system status and the model's computational cost. Therefore, when the agent $\pi$ is fine-tuned using delay prediction, it will be pushed to adopt a better understanding of the host platform's characteristics and the costs of different options of $M$ on that platform.
Moreover, delay times can be directly observed, and can be a \textit{convenient supervision signal on a new platform} since we do not have signals from our main task (which requires labeled video data).
Therefore, on a new platform, we can fine-tune our agent to predict the delay by using the observed delay times as supervision, which will improve the agent's understanding of the device.

Although the delay prediction task improves the quality of the intermediate features (which are also used to select actions), fine-tuning with it does not necessarily lead to a better policy on the host device. 
Hence, to better align the performance on the auxiliary task of delay prediction with policy learning, we further propose a meta-optimization-based learning framework to find a good initialization for the agent, such that updating based on the loss of delay prediction would also improve the quality of the policy. 
Combining the auxiliary task and the meta-learning framework allows us to train SAN on a set of local hardware platforms, and perform only a quick deployment-time adaptation on the target device at run-time, without requiring the actual labeled video dataset on that target device. 

\textbf{Auxiliary task of delay prediction.} 
When SAN is deployed to an unseen target device, the processing times of each action at a given system status will shift significantly due to the drift in computing profiles. Therefore, fine-tuning on the target device is required to adapt SAN to the new hardware characteristics.
Due to the lack of ground-truth labels for the streaming videos at run-time, 
we introduce a self-supervised auxiliary task of delay prediction for adapting SAN to unseen hardware platforms. 
The delay prediction task calibrates the learned system-status-aware features for the agent module $\pi$ using supervisory signals (processing times) that can be \textit{generated on-the-fly} with only a simple timer on the device.

Specifically, an auxiliary task head is added to the agent $\pi$, as shown in Fig.~\ref{fig: pipeline}.
The auxiliary task head takes in the features from the agent as well as the chosen action $a_t$, to predict the delay time $d_t$.
To train our model on the auxiliary task, we apply a mean-squared-error loss between our predictions $\hat{d}_t$ and the observed delay $d_t$. The auxiliary loss $\mathcal{L}_{aux}$ at step $t$, 
is thus deﬁned as:
$\mathcal{L}_{aux,t} = (d_t-\hat{d}_t)^2.$

At deployment time, we fine-tune the agent $\pi$ on the target device using this auxiliary task only, allowing the agent to better understand the target device's characteristics and processing times.
However, this does not necessarily mean that the agent has optimized its policy output -- in order to achieve that, we further introduce a meta-optimization method as described next.

\textbf{Meta-optimization for self-supervised agent adaptation.}
During deployment onto the target device, optimizing the auxiliary task of delay prediction is loosely connected to improving the actual policy. To better bind the quality of the developed policy to the delay prediction performance, we introduce a meta-optimization step to our agent $\pi$ during training, such that fine-tuning of the auxiliary task will also lead to a \textit{better adaptation of the policy}.

We approach this problem by simulating the fine-tuning process using the delay prediction error on multiple target platforms, and (meta) optimize $\phi$ such that the update based on the delay prediction error leads to better adaptation of the policy as well.
This will effectively link the quality of the policy to the delay prediction performance.
Here, we adopt a meta-learning framework to tackle the problem, treating $\phi$ as the \textit{meta-parameters} to meta-optimize.
In this context, we denote the auxiliary task as the \textit{meta-train task} and the agent's policy generation as the \textit{meta-test task}.
Our meta-optimization algorithm aims to learn a good initialization of $\phi$, such that when it is updated via the meta-train task, it can achieve good performance on the meta-test task.

We assume that our SAN is first trained on a source device $D_{src}$, and then adapted to the additional $K$ target platforms $\{D_{tgt}^k \}_{k=1}^K$\footnote{We empirically find that we can simulate a large set of target platforms by setting different utility limitations on a few selected devices, as described in our experiments.}.
After first training our SAN on the source device $D_{src}$, we freeze the main network $M$, and keep a copy of the agent's parameters (denoted as $\phi_{meta,0}$) to initialize the meta-learning process.
In the $u$-th meta-learning iteration, we first meta-train the agent for each target device $D_{tgt}^k$ using its loss on the auxiliary task and perform a pseudo-update on the meta parameters $\phi_{meta,u}$, which can be formalized as:
\begin{equation}
\label{eq:metatrain_update}
\phi_{tgt,u}^k \leftarrow \phi_{meta,u} - \alpha \nabla_{\phi_{meta,u}}\mathcal{L}_{aux,t}(\phi_{meta,u}, D_{tgt}^k)
\end{equation}
where $\phi_{tgt}^k$ refers to the parameters of the meta-trained agent $\pi_{tgt}^k$ which is about to be deployed onto the target device $D_{tgt}^k$, and $\alpha$ is the learning rate for meta-training. 

Then, we check the performance of the pseudo-updated parameters on target device by  meta-testing the performance of the pseudo-updated agent $\pi_{tgt}^k(:;\phi_{tgt,u}^k)$ for the agent's policy generation following:
\begin{equation}
\mathcal{L}_{meta}^k (\phi_{tgt,u}^k) = \mathcal{L}_{\pi}(\phi_{tgt,u}^k, D_{tgt}^k),
\end{equation}
where $\mathcal{L}_{meta}^k$ is the meta-testing-loss from the target
hardware platform $D_{tgt}^k$.
Intuitively, this loss penalizes our agent based on its performance on a target device after a short period of fine-tuning.
Hence, we update our meta parameters $\phi_{meta,u}$ in the $u$-th meta-learning iteration as follows:
\begin{equation}
\begin{aligned}
\phi _{meta,u+1} \leftarrow &  \phi_{meta,u}  - \beta  \sum_{k=1}^K \nabla_{\phi_{meta,u}}\mathcal{L}_{meta}^k(\phi_{tgt,u}^k),
\end{aligned} 
\end{equation}
where $\beta$ is the learning rate for optimizing the meta parameters, and $\phi_{tgt,u}^k$ is obtained via the pseudo-update in Eq.~\ref{eq:metatrain_update}.
Through updating our model with this meta-learning scheme, our agent parameters will converge towards optimal parameters $\phi_{meta,*}$ that are able to effectively adapt to each target device through fine-tuning via the auxiliary task. 

\subsection{Training and Testing}
\label{Sec: Trainingtesting}
As mentioned above, we consider two scenarios: deploying our SAN on a seen or unseen device.

\textbf{Model training and testing for seen device.}
In this scenario, we evaluate SAN when it is trained and tested on the same device, which evaluates the effectiveness of our SAN design described in Sec. \ref{Sec: SAN}.
For model training, we first utilize the dataset's training set to initialize our main network $M$, which is controlled by a random policy (randomly sampling the resolution-depth pair) to improve its accuracy via the loss $\mathcal{L}_M$. 
Then, we train our agent $\pi$ using the loss $\mathcal{L}_{\pi}$ with the parameters of the main network ($M$) fixed.
Lastly, we fine-tune the main network $M$ and agent $\pi$ jointly in a fully end-to-end manner with both $\mathcal{L}_M$ and $\mathcal{L}_{\pi}$, allowing us to obtain a high-performance video understanding dynamic network controlled by an optimal system-status-aware policy on the current local device.
For evaluation, we test the trained SAN on the same device, but with an unseen system status trajectory, which evaluates both the accuracy and the delay of our SAN.

\textbf{Model meta-optimization and deployment for unseen device.} 
In this scenario, our SAN is pre-trained and meta-optimized on a set of seen source devices, and we evaluate its efficacy when deployed to an unseen device. This evaluates the efficacy of our MSA design described in Sec.~\ref{Sec: MSA}.
At the start of training, we first initialize and pre-train our SAN on a seen device, following the process described previously.
Then, during meta-optimization, we freeze the main network $M$ and only optimize our agent $\pi$.
During meta-optimization iterations, we randomly sample video clips from the training set for meta-training and meta-testing.
For better generalization, video clips used for meta-training are not used for meta-testing.
When deploying our SAN to an unseen target device, we first fine-tune the agent $\pi$ via the auxiliary task of delay prediction.
The adapted agent is then tested for average accuracy and delay over all frames in the testing set.
When comparing our method's average accuracy and delay with others, we compute the global accuracy and delay (counting both the fine-tuning and final testing phase) for a fair comparison.

\section{Experiments}
\label{sec: Experiments}
We describe our setup and results below. More analysis and details are provided in Supplementary. 

\textbf{SAN's architecture.} 
For our dynamic encoder, we adopt the fully-convolutional backbone ResNet50 \cite{he2016deep} to handle various input resolutions. We add three exit ports (i.e., $n=3$) at the end of these specific layers of ResNet50 ($\{conv3\_x, conv4\_x, conv5\_x\}$ \cite{he2016deep}) to achieve dynamic depth. 
Moreover, we specifically design a convolutional layer for each resolution-depth pair to unify the shapes of the encoder's output features (details in Supplementary).
Global average pooling is performed on the unified encoder's output and send to the LSTM to update its hidden state $h \in \mathbb{R}^{2048}$.
The output of the LSTM is fed into the task head for online video understanding. 
On the other hand, our lightweight agent $\pi$ consists of three fully connected layers.
We utilize three widely used monitors \cite{nvitop} to collect system information, which contains the status of system's CPU, GPU, memory, disk, I/O and fan. 
For model training, we set $\lambda_{acc} = 2$, $d_b = 0.03$, $\alpha = 1e^{-3}$ and $\beta = 1e^{-5}$. More details of our SAN are presented in Supplementary. 

\textbf{Online video task settings.} We compare the performance of our SAN with the prior methods on two widely used online video understanding tasks: 
1) \textbf{Online action recognition:} 
We follow the previous work \cite{perrett2017recurrent} to use the 50Salads dataset \cite{stein2013combining}, since it contains very long videos that are suitable for online video action recognition.
Here, we consider three (i.e., $m=3$) resolution candidates:~$[112,168,224]$ and use a fully-connected layer as the task head.
We compare our method with SOTA online action recognition methods, including RA \cite{perrett2017recurrent}, DDLSTM \cite{perrett2019ddlstm}, AR-Net \cite{meng2020arnet}, and LiteEval \cite{wu2019liteeval}. 
To test models on the same platform for a fair comparison, we use their publicly released code (AR-Net) when available and implement the rest (RA, DDLSTM and LiteEval) on our own.
2) \textbf{Online pose estimation: } 
We follow the existing video pose estimation works \cite{fan2021motion,habibian2021skip} to conduct our experiments on the Sub-JHMDB dataset \cite{jhuang2013towards}, a collection of 11,200 frames from 316 video clips, labeled with 15 body joints. 
Here, we consider three (i.e., $m=3$) resolution candidates:~$[128,192,256]$ and follow the work of \cite{xiao2018simple} to use 3 deconvolutional layers to build the task head for pose estimation. 
Moreover, 
we replace LSTM with a ConvLSTM \cite{shi2015convolutional} to accommodate for 2D feature maps.
We compare our method with the existing dynamic pose estimation networks, i.e., DKD \cite{nie2019dynamic} and Skip-Convolution \cite{habibian2021skip}.
Similarly, we re-implement DKD and Skip-Convolutions as there is no publicly available code.

\begin{table}
\caption{
Results on 50Salads dataset for online action recognition.
}
\vspace{-0.3cm}
\label{Tab: 50Salads Result}
\resizebox{0.477\textwidth}{!}{%
\begin{tabular}{|l|cccc|}
\hline
Method & Accuracy & Max Delay & Mean Delay & R\textbackslash T Frames \\ \hline
RA \cite{perrett2017recurrent} & 42.6 \% & 76.6 $ms$ & 33.9 $ms$ & 79.1 \%\\
DDLSTM \cite{perrett2019ddlstm} & 41.1 \% & 407.4 $ms$ & 110.3 $ms$ & 5.9 \%\\
LiteEval \cite{wu2019liteeval} & 40.3 \% & 110.7 $ms$ & 68.9 $ms$ & 30.4 \%\\
AR-Net \cite{meng2020arnet} & 40.9 \% & 91.1 $ms$ & 56.2 $ms$ & 62.1\% \\ \hline
{main network} &  &  &  & \\ 
+ random policy & 46.2\% & 80.5 $ms$ & 34.1 $ms$ & 92.1\%\\
+ stream-aware & \textbf{55.4} \% & 80.8 $ms$ & 40.3 $ms$ & 82.0\%\\
+ SAN  & 53.8\% & \textbf{49.4} $ms$ & \textbf{29.9} $ms$ & \textbf{95.4}\% \\ \hline
\end{tabular}
}%
\vspace{-0.3cm}
\end{table}

\begin{table*}[t]
\centering
\caption{Results on Sub-JHMDB dataset for online pose estimation.
}
\vspace{-0.3cm}
\label{Tab: JHMDB Result}
\resizebox{1\textwidth}{!}{%
\begin{tabular}{|l|cccccccc|ccc|}
\hline
Method & Head & Sho. & Elb. & Wri. & Hip & Knee & Ank. & Avg & Max Delay & Mean Delay & R\textbackslash T Frames \\ \hline
DKD \cite{nie2019dynamic} & 98.3 & 96.6 & 90.4 & 87.1 & 99.1 & 96.0 & 92,9 & 94.0 & 133.8 $ms$ & 47.9 $ms$ & 59.9 \% \\
SimpleBaseline \cite{xiao2018simple} & 97.5 & 97.8 & 91.1 & 86.0 & 99.6 & 96.8 & 92.6 & 94.4 & 129.6 $ms$ &  45.5 $ms$ & 63.8 \% \\
Skip-Convolution \cite{habibian2021skip} & 98.7 & 97.7 & 92.0 & 88.1 & \textbf{99.3} & \textbf{96.6} & \textbf{91.0} & 95.1 & 115.7 $ms$ & 51.2 $ms$ & 53.7 \% \\\hline
main network + random policy & 97.5 & 95.2 & 91.4 & 93.5 & 77.7 & 86.8 & 72.4 & 86.9 & 130.9 $ms$ & 36.1 $ms$ & 72.9\%\\
main network + stream-aware & \textbf{99.4} & \textbf{98.6} & \textbf{97.5} & \textbf{98.1} & 93.9 & 94.2 & 90.0 & \textbf{95.3} & 126.1 $ms$ & 44.7 $ms$ & 63.7\% \\
main network + SAN & 99.1 & 98.7 & 97.3 & 98.2 & 90.1 & 92.5 & 85.0 & 93.4 & \textbf{41.2 $ms$} & \textbf{33.1 $ms$} & \textbf{81.4\%} \\ \hline
\end{tabular}
}%
\vspace{-4mm}
\end{table*}

\begin{figure}[t]
\centering
\begin{minipage}[t]{0.5\linewidth}
\raggedleft
\includegraphics[width=1\linewidth]{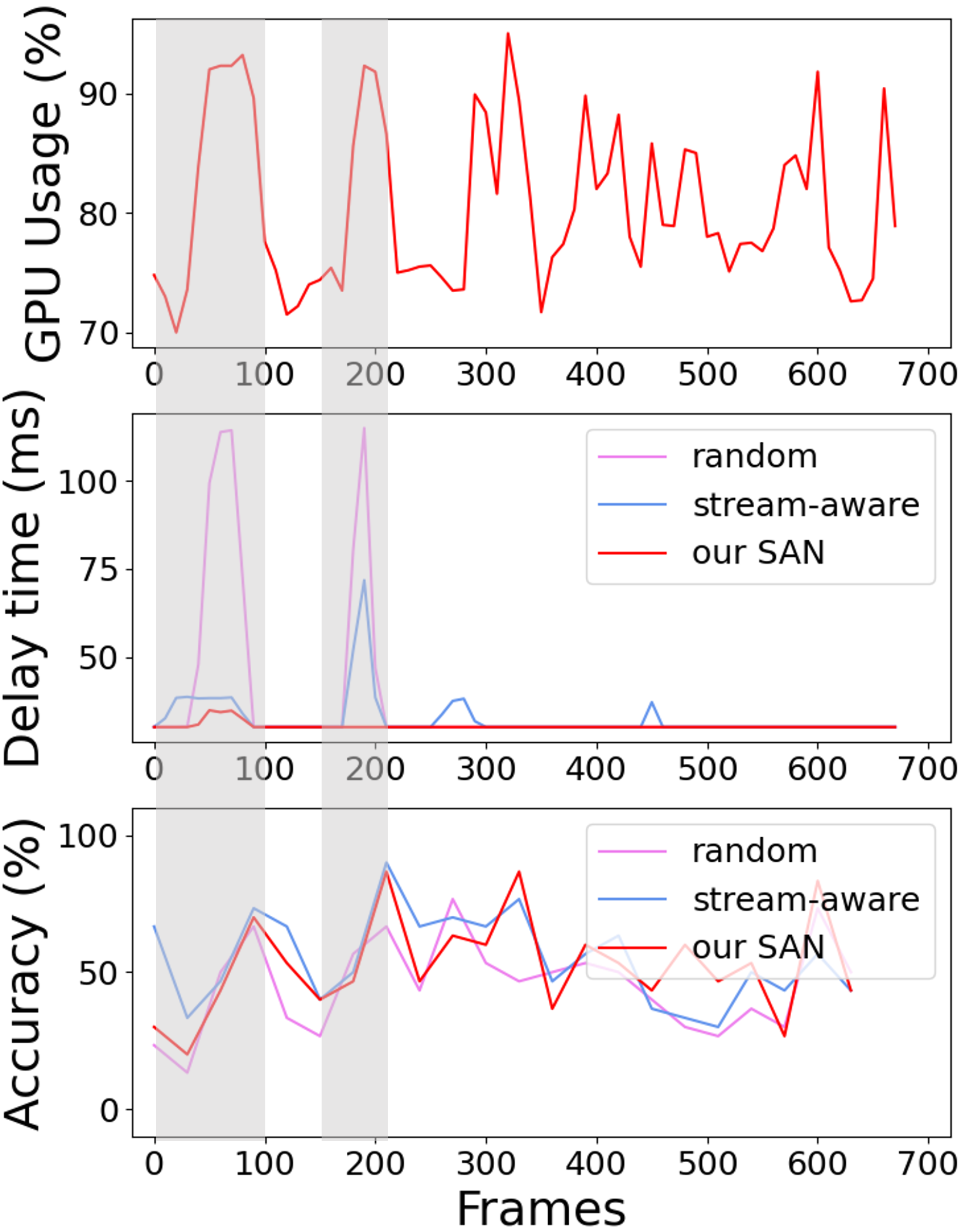}
%\caption{fig2}
\end{minipage}%
\begin{minipage}[t]{0.5\linewidth}
\raggedleft
\includegraphics[width=1\linewidth]{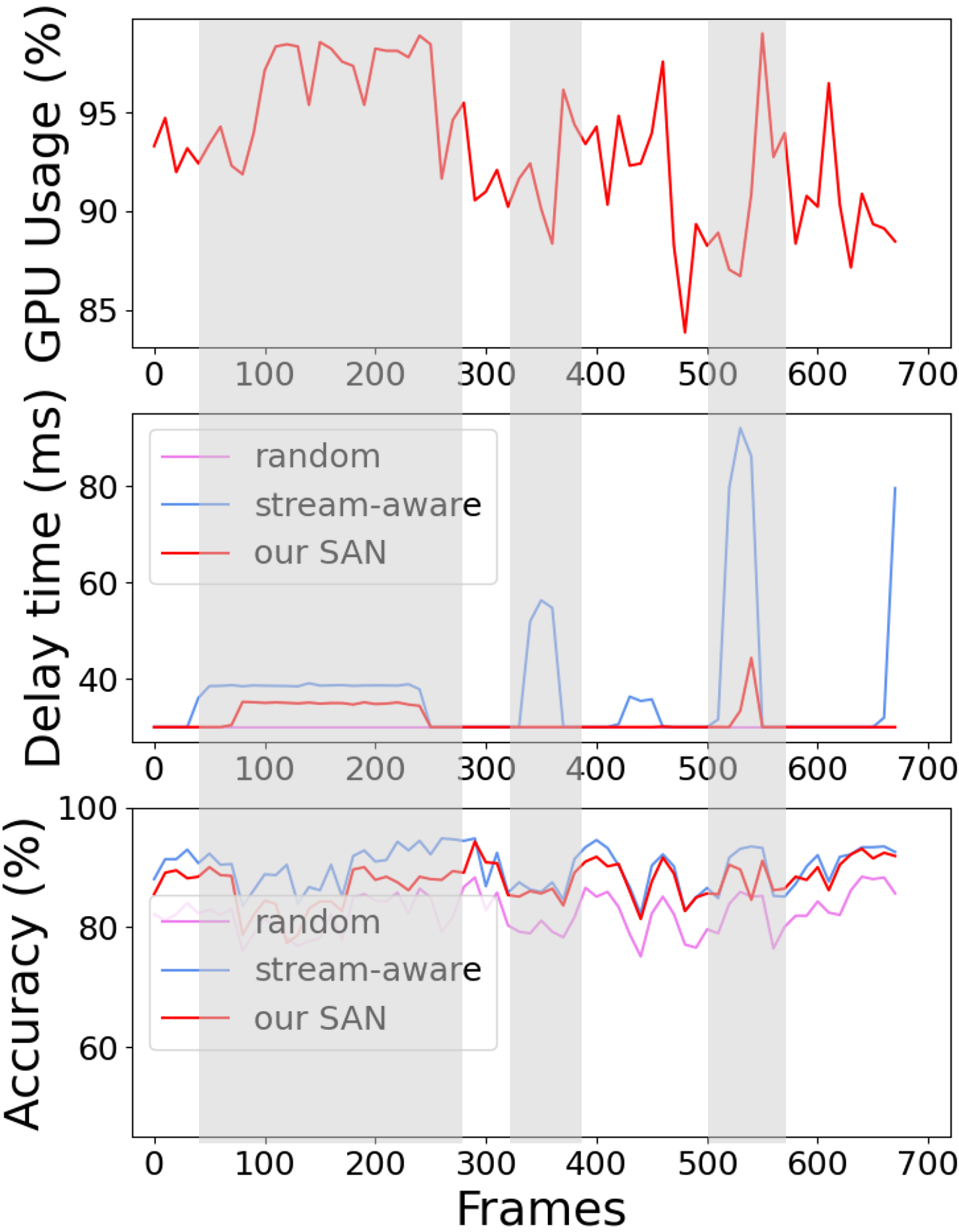}
%\caption{fig2}
\end{minipage}%
\vspace{-0.3cm}
\caption{A trace of models' accuracy and delay for online video-based action recognition (left) and pose estimation (right). In the areas masked by the gray color, we can observe a rapid increase of the baseline model's delay (see the second row), which corresponds to the high-load status of the system. 
}
\label{Fig: dynamic system}
\vspace{-4mm}
\end{figure}

\textbf{Experiment settings.} To build \emph{dynamic systems with varying system status}, we design multiple background processes 
(i.e., matrix calculators, video compressors, and large deep learning models) 
to occupy various amount of computational resources.
We can then generate dynamic system load trajectories using these processes to simulate dynamic system load environments.
During model training, we randomly generate dynamic system trajectories in each iteration  to simulate various dynamic environments to train our SAN. During  testing, for fair comparisons, we first generate 3 dynamic trajectories unseen in training, and then run all the evaluation experiments on these 3 fixed trajectories and report the average result. 
To evaluate the models' efficiency, we measure the maximum delay and mean delay for each model. Moreover, as the videos of both datasets \cite{stein2013combining,jhuang2013towards} are at 30 fps, we consider delay time $<30$ $ms$ as \emph{real-time}, and measure the percentage of such real-time predictions (R\textbackslash T Frames).

\subsection{Evaluating SAN on seen devices}
\label{Sec: SAN Result}

First, we compare our SAN with the SOTA methods for both tasks and report the results in Tab.~\ref{Tab: 50Salads Result} and Tab.~\ref{Tab: JHMDB Result}. 
In both tables, we also report the performance of two baselines of our method: \textbf{main network + random policy} where actions are randomly selected; and \textbf{main network + stream-aware} where the agent $\pi$ uses information from the input stream without considering system status.
Both baselines are not system-status-aware.

\textbf{Online action recognition.} Tab.~\ref{Tab: 50Salads Result} presents the results on the 50Salads dataset for online action recognition. We compare our method with four SOTA methods, including two static models (i.e., RA \cite{perrett2017recurrent} and DDLSTM \cite{perrett2019ddlstm}) and two efficient dynamic models (i.e., AR-Net \cite{meng2020arnet} and LiteEval \cite{wu2019liteeval}). 
Our SAN outperforms these methods on all metrics, achieving a higher accuracy while maintaining a low mean delay (29.9ms) and high percentage of real-time frames (95.4\%).
These improvements are due to our system-status-aware design of SAN, allowing it to make on-the-fly adaptations to process a higher percentage of frames in real-time.
The two baselines are not system-status-aware and therefore perform significantly worse on the delay-related metrics, especially on max delay (80.5ms and 80.8ms vs our 49.4ms). Compared to the random policy, stream-aware policy significantly improves the accuracy at the cost of reduced real-time performance. 
Overall, our SAN achieves superior performance and at the same time also decreases the delay.

\textbf{Online pose estimation.} Tab. \ref{Tab: JHMDB Result} presents the results on the Sub-JHMDB dataset for online pose estimation. We compare our SAN with a static network SimpleBaseline \cite{xiao2018simple} and two dynamic efficient networks: DKD \cite{nie2019dynamic} and Skip-Convolution \cite{habibian2021skip}. 
As shown in Tab. \ref{Tab: JHMDB Result}, our main network + stream-aware baseline can achieve the SOTA performance but may still incur large delays occasionally. 
With the help of our system-status-aware design, our SAN can control the maximum and mean delay at a low level while still maintaining high accuracy.

\begin{table*}[ht]
\caption{
Results of policy adaptation across platforms for online pose estimation. More results on other platforms are in Supplementary. 
}
\vspace{-0.3cm}
\label{Tab: MSA}
\resizebox{1\textwidth}{!}{%
\begin{tabular}{|l|ccc|ccc|ccc|}
\hline
\multirow{2}{*}{Method} & \multicolumn{3}{c|}{a+b $\rightarrow$ c} & \multicolumn{3}{c|}{a+c $\rightarrow$ b} & \multicolumn{3}{c|}{b+c $\rightarrow$ a} \\ \cline{2-10} 
& Accuracy & Mean Delay & R\textbackslash{}T Frames & Accuracy & Mean Delay & R\textbackslash{}T Frames & Accuracy & Mean Delay & R\textbackslash{}T Frames \\ \hline
Fully-supervised learning (Upper bound)  & 93.0 \% & 28.3 $ms$ & 87.4 \% & 93.1 \% & 24.7 $ms$ & 84.9 \% & 92.8 \% & 28.1 $ms$ & 74.5 \%\\ \hline
Direct transfer & 87.3 \% & 28.4 $ms$ & 82.4 \%  & 85.7 \% & 25.0 $ms$ & 79.5 \% & 90.9 \% & 33.5 $ms$ & 20.4\%  \\
Feature alignment & 90.8 \% & 29.0 $ms$ & 80.4 \% & 91.0 \% & 27.2 $ms$ & 74.6 \% & 91.0 \% & 31.8 $ms$  & 64.5 \% \\\hline
Our self-supervised fine-tune & 91.1 \% & 29.8 $ms$  & 79.8 \% & 85.8 \% & 21.9 $ms$ & 84.1 \% & 91.5 \% & 29.3 $ms$ & 61.7 \% \\
Our MSA & 92.8 \% & 29.7 $ms$ & 86.1 \%  & 92.9 \% & 25.1 $ms$ & 82.2 \% & 92.4 \% & 30.0 $ms$ & 71.9 \% \\ \hline
\end{tabular}
}%
\vspace{-4mm}
\end{table*}

\begin{table*}[ht]
\caption{
Results of policy adaptation across platforms for online action recognition. 
}
\vspace{-0.3cm}
\label{Tab: MSA_AR}
\resizebox{1\textwidth}{!}{%
\begin{tabular}{|l|ccc|ccc|ccc|}
\hline
\multirow{2}{*}{Method} & \multicolumn{3}{c|}{a+b $\rightarrow$ c} & \multicolumn{3}{c|}{a+c $\rightarrow$ b} & \multicolumn{3}{c|}{b+c $\rightarrow$ a} \\ \cline{2-10} 
& Accuracy & Mean Delay & R\textbackslash{}T Frames & Accuracy & Mean Delay & R\textbackslash{}T Frames & Accuracy & Mean Delay & R\textbackslash{}T Frames \\ \hline
Fully-supervised learning (Upper bound)  & 53.8 \% & 26.3 $ms$ & 95.4 \% & 52.6 \% & 31.3 $ms$ & 93.1 \% & 52.1 \% & 31.9 $ms$ & 90.4 \%\\ \hline
Direct transfer & 41.7 \% & 28.1 $ms$ & 77.3 \%  & 47.2 \%& 34.6 $ms$ & 84.7 \% & 52.0 \%& 41.1 $ms$ & 40.6 \%  \\
Feature alignment   & 50.1 \%& 29.8 $ms$ & 82.4 \% & 50.3 \%& 32.7 $ms$ & 88.1 \% & 51.7 \%& 37.2 $ms$ & 60.5 \%\\ \hline
Our self-supervised fine-tune &  49.1 \%& 26.5 $ms$ & 93.8 \%  & 51.0 \%& 32.2 $ms$ & 93.1 \% & 50.7 \%& 33.7 $ms$ & 87.2 \%  \\
Our MSA & 52.6 \%& 26.8 $ms$ & 94.1 \% & 52.2 \%& 31.6 $ms$ & 92.7 \% & 51.9 \%& 32.1 $ms$ & 90.2 \%\\ \hline
\end{tabular}
}%
\vspace{-0.3cm}
\end{table*}

\textbf{Qualitative analysis.}
We analyze the effect of the system load dynamics and show results from test-time runs on a laptop with a NVIDIA Geforce GTX 1080 GPU (11 GB), where 10 background processes are randomly activated.
Fig. \ref{Fig: dynamic system} presents the models' frame-level delay and accuracy along with the system status. We compare our SAN with two baselines: \textbf{main network + stream-aware} and \textbf{main network + random policy}.
In the first row of Fig. \ref{Fig: dynamic system}, we observe that the GPU load fluctuates over time, varying the available computational resources for the online models.
As a result, the delay of models with random/stream-aware policy fluctuates up to around 100ms as shown in the second row. 
In contrast, by using the system-status information to control the model, our SAN is able to stabilize the delay and avoid large delays amid the dynamic system. 
Moreover, as shown in the last row, with the help of the stream-aware design, our SAN can still obtain a similar accuracy to the stream-aware model.
These results show the advantage of our system-status-aware and stream-aware designs, which can simultaneously control the real-time delay and attain high accuracy for online video understanding tasks.

\subsection{Evaluating SAN with MSA on unseen devices}
\label{Sec: MSA Result}
Our proposed MSA is introduced to achieve effective adaptation to the unseen host devices. Here, we compare our MSA with three baselines: 1) \textbf{Direct transfer}: Directly deploying our SAN to the unseen device. 2) \textbf{Feature alignment}: Minimizing the domain gap between the source device and target device via a domain discriminator \cite{ganin2016domain,Mitsuzumi_2021_CVPR}.
3) \textbf{Self-supervised fine-tune}: Using our auxiliary task (delay prediction) to fine tune the model without meta optimization.
To further demonstrate the efficacy of MSA, we also present the upper bound, dubbed as \textbf{Fully-supervised learning}, where we directly train the agent 
on the target environment, which eliminates the need for adaptation. 

Here, we collect three hardware platforms for training and testing: 
1) \textbf{Device a:} a laptop with an AMD Ryzen Threadripper 2950X and an NVIDIA Geforce GTX 1080 GPU (11 GB).
2) \textbf{Device b:} a desktop with an Intel(R) Core(TM) i9-10900 and an NVIDIA Geforce RTX 2080 GPU (12 GB).
3) \textbf{Device c:} a workstation with an AMD Ryzen Threadripper 3960X and an NVIDIA Geforce RTX 3090 GPU (24 GB).
We randomly sample two platforms as the seen source devices for model initialization and meta-optimization and use the remaining one as the unseen target device for model adaptation (e.g., $Source:\{a,b\}\rightarrow Target:\{c\}$). Furthermore, we augment the source platforms by restricting the number of available CPU and the highest frequencies of CPU and GPU to achieve various  environments with different resource configurations for better training. The results of policy adaptation are reported in Tab.~\ref{Tab: MSA} and \ref{Tab: MSA_AR}. More results on other portable devices (e.g., NVIDIA Jetson TX2) are presented in Supplementary.

As shown in Tab.~\ref{Tab: MSA} and \ref{Tab: MSA_AR}, the direct transfer method fails to maintain the high accuracy and low delay simultaneously due to the hardware configuration shift between devices. 
On the other hand, by aligning the system-status-aware features between the source and target devices, the agent finds a sub-optimal policy to balance the accuracy and delay.
Using the auxiliary task to fine-tune the agent without meta optimization can achieve real-time predictions on the target device, but it sometimes fails to maintain high accuracy.
Finally, our MSA simultaneously maintains a high ratio of real-time predictions and accurate predictions. Furthermore, the results are close to the upper bound of actually training on the target device, showing the effectiveness of our MSA to generalize to unseen devices.

\subsection{Ablation Study}
\label{Sec: additional experiment}

We further analyze the characteristics of our dynamic resolution and dynamic depth mechanism, and investigate the performance of our agent under various delay thresholds $d_b$ and accuracy coefficients $\lambda_{acc}$.

First, we evaluate the impact of the dynamic resolution and depth mechanism and perform ablation studies by allowing one of them to be dynamic while fixing the other. 
In Tab.~\ref{Tab: ablation study dynamic machanism}, we observe that, by allowing for dynamic resolutions or dynamic depths, our main network can reduce the maximum and mean delay while maintaining high accuracy.
Moreover, by combining these two mechanisms, our SAN can further reduce the mean delay and suppress the maximum delay at a lower level, with even better accuracy.

Next, we investigate the performance of our SAN under different delay thresholds $d_b$ and accuracy coefficients $\lambda_{acc}$. 
As shown in Tab. \ref{Tab: AB_hyber}, under different values of $d_b$, our SAN can still significantly decrease the maximum delay while maintaining high accuracy. 
Furthermore, as $d_b$ increases, SAN is allowed to have higher delays before getting penalized, leading to higher accuracy while slightly degrading delay metrics, which is what we expected.
We also evaluate different values of $\lambda_{acc}$ and present the results in Tab. \ref{Tab: AB_hyber}. 
We observe that our SAN generally achieves good accuracy and delay performance for different values of $\lambda_{acc}$, showing that it is robust to the $\lambda_{acc}$ hyperparameter.
Also, when $\lambda_{acc}$ increases, the accuracy improves and delay slightly degrades, as we would expect.

\begin{table}[t]
\centering
\caption{Ablation results for dynamic mechanism.
}
\vspace{-3mm}
\label{Tab: ablation study dynamic machanism}
\resizebox{1\columnwidth}{!}{%
\begin{tabular}{|l|ccc|}
\hline
Setting & Avg & Max Delay & Mean Delay \\ \hline
$res=256,dep=3$ & 94.4 \% & \ 129.6 $ms$ & \ 45.5 $ms$ \\ \hline
$res=256,dep=[1,3]$ & 92.6 \% & 71.4 $ms$ & 39.3 $ms$ \\
$res=256,dep=[1,2,3]$ & 93.3 \%& 63.6 $ms$ & 34.5 $ms$ \\\hline
$res=[128,256],dep=3$ & 91.2 \%& 84.0 $ms$ & 43.7 $ms$ \\
$res=[64,128,192,256],dep=3$ & 91.9 \%& 61.9 $ms$ & 38.0 $ms$\\\hline
Our SAN & 93.4 \%& 41.2 $ms$ & 33.1 $ms$\\\hline
\end{tabular}
}%
\vspace{-3mm}
\end{table}

\begin{table}[t]
\centering
\tabcolsep=6mm
\caption{
Ablation results for different values of $d_b$ and $\lambda_{acc}$.
}
\vspace{-3mm}
\label{Tab: AB_hyber}
\resizebox{1\columnwidth}{!}{%
\begin{tabular}{|l|ccc|}
\hline
Setting & Avg & Max Delay & Mean Delay \\ \hline
$d_b = 10\,\,ms$ & 92.4 \% & 39.2 $ms$ & 27.7 $ms$  \\
$\,\,\,\,\,\,\, = 20\,\,ms$ & 92.8 \% & 39.4 $ms$ & 30.7 $ms$ \\ 
$\,\,\,\,\,\,\, = 30\,\,ms$ & 93.4 \% & 41.2 $ms$ & 33.1 $ms$ \\ 
$\,\,\,\,\,\,\, = 40\,\,ms$ & 94.0 \% & 43.5 $ms$ & 41.9 $ms$  \\ \hline

$\lambda _{acc} = 1$ & 91.5 \%  & 41.0 $ms$ & 30.8 $ms$\\
$\,\,\,\,\,\,\,\,\,\,\,\,\, = 2$  & 93.4 \%  & 41.2 $ms$ & 33.1 $ms$\\
$\,\,\,\,\,\,\,\,\,\,\,\,\, = 4$  & 94.0 \%  & 44.6 $ms$ & 37.1 $ms$\\
\hline
\end{tabular}
}%
\vspace{-5mm}
\end{table}

\section{Conclusion}
In this work, we tackle the efficiency problem of deep networks on devices with computational ﬂuctuations. To the best of our knowledge, this is the ﬁrst time this problem is being explicitly tackled for online video understanding tasks. 
We propose SAN, a novel system-status-aware adaptive network for low-delay online action recognition and online pose estimation tasks with a fluctuating computation budget. 
A self-supervised meta optimization framework (MSA) is also proposed for more effective adaptation between hardware platforms. 
Experiments show that our proposed SAN and MSA obtains SOTA performance with low delay even under a rapidly fluctuating system load.

\noindent
\textbf{Acknowledgments.}
This work is supported by MOE AcRF Tier 2 (Proposal ID: T2EP20222-0035), National Research Foundation Singapore under its AI Singapore Programme (AISG-100E-2020-065), and SUTD SKI Project (SKI 2021\_02\_06).

%%%%%%%%% REFERENCES
{\small
\bibliographystyle{ieee_fullname}
\bibliography{egbib}
}

\clearpage

%%%%%%%% Supplementary
%%when combining, remember to clear the footnotes
\setlength{\textheight}{8.875in}
\setlength{\textwidth}{6.875in}
\setlength{\columnsep}{0.3125in}
\setlength{\topmargin}{0in}
\setlength{\headheight}{0in}
\setlength{\headsep}{0in}
\setlength{\parindent}{1pc}
\setlength{\oddsidemargin}{-.304in}
\setlength{\evensidemargin}{-.304in}

%%%%%%%%% TITLE
\title{\vspace{-0.2em} \Large \textbf{System-status-aware Adaptive Network for Online Streaming Video Understanding (Supplementary Material)} \vspace{0.3em} }

\author{Lin Geng Foo\textsuperscript{1\dag}
% \thanks{ \dag~Equal contribution;~~\S~Currently at Meta;~~\ddag~Corresponding author}
~~~ Jia Gong\textsuperscript{1\dag}
~~~ Zhipeng Fan\textsuperscript{2\S}
~~~ Jun Liu\textsuperscript{1\ddag} \\
\textsuperscript{1}Singapore University of Technology and Design ~~~~ \textsuperscript{2}New York University\\
{\tt\small \{lingeng\_foo,jia\_gong\}@mymail.sutd.edu.sg, zf606@nyu.edu, jun\_liu@sutd.edu.sg } \\
}

\maketitle

\setcounter{section}{0}
\setcounter{table}{0}
\setcounter{figure}{0}
\setcounter{equation}{0}

In this Supplementary, we first introduce more details of our experimental setup in Sec.~\ref{Sec: dynamic system setting}.
Then in Sec.~\ref{Sec: more other results}, we present more experiments, including performance on other platforms in Sec.~\ref{Sec: more results on various platforms.} and additional analysis in Sec.~\ref{Sec: additional analysis}.
Next, we describe more details of our dynamic main network and the agent module in Sec.~\ref{Sec: SAN architecture}.
Finally, we have some further discussions regarding related work in Sec.~\ref{Sec: related work}.

\section{More Details of Experimental Setup}
\label{Sec: dynamic system setting}

\begin{figure}[ht]
\centering
\begin{minipage}[t]{0.95\linewidth}
\raggedleft
\includegraphics[width=1\linewidth]{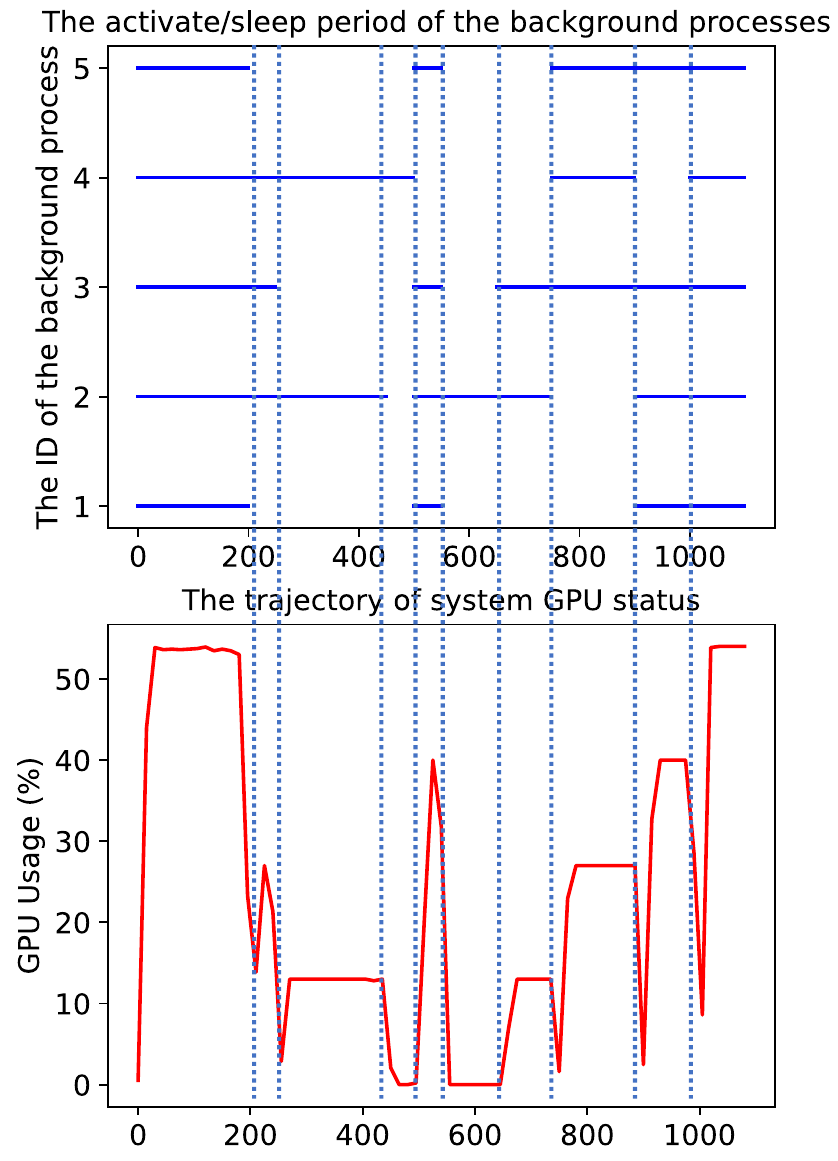}
\end{minipage}%
\caption{
A schedule of 5 background process' activation/sleep periods (top) with the corresponding system status trajectory (bottom).
Note that, in our background process schedule (top), the $y$ axis corresponds to the ID of the background process (which ranges from 1 to 5 in this figure), and a solid line means that the specific process is activated for the corresponding duration indicated on the $x$ axis. 
For the sake of simplicity, in the bottom plot, we only plot the GPU Usage rates, which is a part of our system status $sys_t$.
The GPU Usage rates fluctuate up and down corresponding to the activated background processes, which shows that we can generate different system status trajectories by activating or deactivating background processes. 
}
\label{Fig: trajectory generation}
\end{figure}

Here, we present the details of how we build the dynamic system and test the online models. We use several computation programs (e.g., matrix calculation and deep model inference) to act as background processes to occupy computation resources, which leads to a dynamic and fluctuating system status.

Specifically, to generate various system statuses via background processes, we prepare three kinds of controllable programs: 
1) Matrix calculation, where three types of matrix operations are used in our experiments, e.g., matrix addition/subtraction, matrix scalar multiplication, and matrix multiplication. The matrix size varies from $[4,64,64]$ to $[16,512,512]$.
2) Video compression, where we use python to call FFmpeg \citeSupp{ffmpeg}, to convert videos to MP4 format.
3) Deep model inference, where we run various deep learning models, including ResNet101 \citeSupp{he2016deep2}, HrNet-w32 \citeSupp{sun2019deep} and U-Net \citeSupp{RFB15a}. 
Note that the same program can have multiple parallel processes, e.g. multiple matrix calculations can be performed simultaneously to process different matrices.

We write a script to control the above-mentioned background processes to generate the dynamic system status.
By activating different background processes at different time steps according to a generated schedule, we can create various system status conditions, as shown in Fig.~\ref{Fig: trajectory generation}.
During training, we randomly generate various schedules (with each schedule using different processes with different activation and sleep periods) for training the models. 
For testing, we first randomly generate three schedules with dynamic system statuses, which have not been seen during training. 
These three sets of background process schedules (with each leading to a different system status trajectory) are used to test the performance of all models, and the final performance is obtained by averaging the results of each model on all three sets.
Notably, we observe that, by controlling the background processes via our script, we can get nearly the same system status trajectory in multiple runs (for a given schedule), and since we average over multiple sets of trajectories, this allows us to have fair comparisons between different models.

\textbf{Implementation Details.}
In Sec. 4.1 of the main paper, we conduct our experiments using a laptop consisting of an AMD Ryzen Threadripper 2950X CPU and an NVIDIA Geforce GTX 1080 GPU (11 GB). On this platform, we prepare four matrix calculators, four deep learning models, and two video compressors as the background processes to generate a dynamic system for online models' evaluation.
The activate/sleep period length of each process has a duration of $[30, 300]$ frames.
In Sec. 4.2 of the main paper, similarly, we set up 10 background processes for all devices. 
In order to sweep the full range of system status in each device, the configurations of the processes on each device are different.

\section{Additional Experiments}
\label{Sec: more other results}

\subsection{Additional analysis}
\label{Sec: additional analysis}

\textbf{The performance of each resolution-depth pair.}
To further observe and analyze the behavior of our learned policy, we first report the performance of each subnetwork (i.e., resolution-depth pair), by testing each individual ``static" subnetwork for online action recognition on 50Salads. 
We present the results in Tab.~\ref{tab: results of each branch}.
It shows that all subnetworks can achieve reasonable performance on online action recognition task with different computational complexity. 

\begin{table}[ht]
\centering
\caption{
Results of each resolution-depth pair for online action recognition.
}
\label{tab: results of each branch}
\resizebox{1\columnwidth}{!}{%
\begin{tabular}{|c|ccc|}
\hline
Resolution-Depth Pair & Acc. (\%) & Max Delay (ms) & Mean Delay (ms) \\ \hline
\{112,1\} & 36.6 & 31.1 & 18.3 \\
\{112,2\} & 39.6 & 42.7 & 24.8 \\ 
\{112,3\} & 45.2 & 66.5 & 42.1 \\
\{168,1\} & 41.9 & 45.5 & 27.2 \\
\{168,2\} & 47.7 & 77.3 & 47.8 \\
\{168,3\} & 54.3 & 98.7 & 70.6 \\
\{224,1\} & 43.6 & 59.3 & 40.5 \\
\{224,2\} & 55.1 & 101.9 & 70.6 \\
\{224,3\} & 56.0 & 174.2 & 125.3 \\ 
\hline
Our SAN   & 53.8 & 49.4 & 29.9 \\
\hline
\end{tabular}
}%
\end{table}

\textbf{The selection frequency of each resolution-depth pair.}
Then, we record the frequency that each subnetwork (i.e., resolution-depth pair) is selected by our agent during the testing phase on the online action recognition task, and present the results in Tab.~\ref{tab: distribution of each branch}.
We observe that each resolution-depth pair is selected at least occasionally, which shows their usefulness, and also the overall efficacy of our design with dynamic resolution and dynamic depth.

\begin{table}[ht]
\caption{
Selection frequency of each subnetwork (i.e., resolution-depth pair) by our agent for online action recognition.
}
\label{tab: distribution of each branch}
\resizebox{1\columnwidth}{!}{%
\begin{tabular}{|l|ccccc|}
\hline
Subnetwork: & \{112,1\} & \{112,2\} & \{112,3\} & \{168,1\} & \{168,2\} \\ \hline
Frequency: & 21.4 \% & 16.7 \% & 2.5 \% & 19.8 \% & 26.0 \% \\
\hline
\hline
Subnetwork: & \{168,3\} & \{224,1\} & \{224,2\} & \{224,3\} & \\ \hline
Frequency: & 3.5 \% & 1.6 \% & 2.8 \% & 5.7 \% & \\\hline
\end{tabular}
}%
\end{table}

\subsection{More results on other portable devices}
\label{Sec: more results on various platforms.}
To further investigate the effectiveness of our SAN and MSA,
we additionally set up experiments on an NVIDIA Jetson TX2 platform with an ARM-based Cortex-A57 CPU and 256-core NVIDIA Pascal GPU (\textbf{device d}), 
a Windows laptop with an Intel Core i5-11400H CPU (\textbf{device e}), 
and an ARM-based M1 macbook (\textbf{device f}).
These experiments are conducted on the 50Salads dataset, and the results are reported in Tab.~\ref{tab: more reuslts of our SAN.} and Tab.~\ref{tab: more reuslts of policy adaptation.}.
Tab.~\ref{tab: more reuslts of our SAN.} shows that, on these platforms, our SAN still obviously outperforms (main network + random policy) on all metrics. It also achieves comparable accuracy to (main network + stream aware), yet the delay of our method is much lower.
Note that (main network + stream aware) is similar to our method (SAN), and the only difference is that the system status is not considered in (main network + stream aware).
Tab. \ref{tab: more reuslts of policy adaptation.} reports the result of adapting the agent pre-trained on two x86 platforms (Device a and Device b as described in the paper) to an x86 unseen device (Device e) and two ARM-based unseen devices (Device d and f).
Results show that our MSA algorithm significantly improves the accuracy from the baseline (Direct Transfer) on all metrics. 

\begin{table*}[ht]
\centering
\caption{Results of our SAN on other platforms (\textbf{device d, e and f}).}
\label{tab: more reuslts of our SAN.}
\resizebox{1\linewidth}{!}{%
\begin{tabular}{|c|ccc|ccc|ccc|}
\hline
\multirow{2}{*}{Method} & \multicolumn{3}{c|}{d} & \multicolumn{3}{c|}{e} & \multicolumn{3}{c|}{f} \\ \cline{2-10} 
 & Accuracy & Max delay & Mean Delay & Accuracy & Max delay & Mean Delay & Accuracy & Max delay & Mean Delay \\ \hline
main network + random policy & 47.1 \% & 251.7 ms & 93.3 ms & 45.9 \% & 174.6 ms & 78.6 ms& 46.5 \% & 177.0 ms & 82.2 ms \\
main network + stream aware & 55.4 \% & 261.3 ms & 112.6 ms & 55.4 \% & 161.9 ms & 82.0 ms& 55.4 \% & 194.7 ms & 110.4 ms\\
main network + our SAN & 54.1 \% & 89.7 ms & 60.4 ms & 55.1 \% & 67.5 ms & 59.1 ms & 54.6 \% & 91.3 ms & 73.4 ms\\ \hline
\end{tabular}
}%
\end{table*}

\begin{table*}[ht]
\centering
\caption{Results of our SAN on other platforms (\textbf{device d, e and f}).}
\label{tab: more reuslts of policy adaptation.}
\resizebox{1\linewidth}{!}{%
\begin{tabular}{|l|ccc|ccc|ccc|}
\hline
\multirow{2}{*}{Method} & \multicolumn{3}{c|}{a+b $\rightarrow$ d} & \multicolumn{3}{c|}{a+b $\rightarrow$ e} & \multicolumn{3}{c|}{a+b $\rightarrow$ f} \\ \cline{2-10} 
& Accuracy & Max Delay & Mean Delay & Accuracy & Max Delay & Mean Delay & Accuracy & Max Delay & Mean Delay\\ \hline
Direct Transfer & 41.5 \% & 94.7 ms & 79.0 ms & 37.9 \% & 85.9 ms & 59.4 ms & 46.8 \% & 114.3 ms & 89.5 ms \\
Our MSA & 51.2 \% & 91.5 ms & 68.4 ms & 50.2 \% & 82.3 ms & 55.1 ms & 53.4 \% & 95.8 ms & 76.3 ms \\ \hline
\end{tabular}
}%
\end{table*}

\begin{figure*}[ht]
\centering
\begin{minipage}[t]{0.8\linewidth}
\raggedleft
\includegraphics[width=1\linewidth]{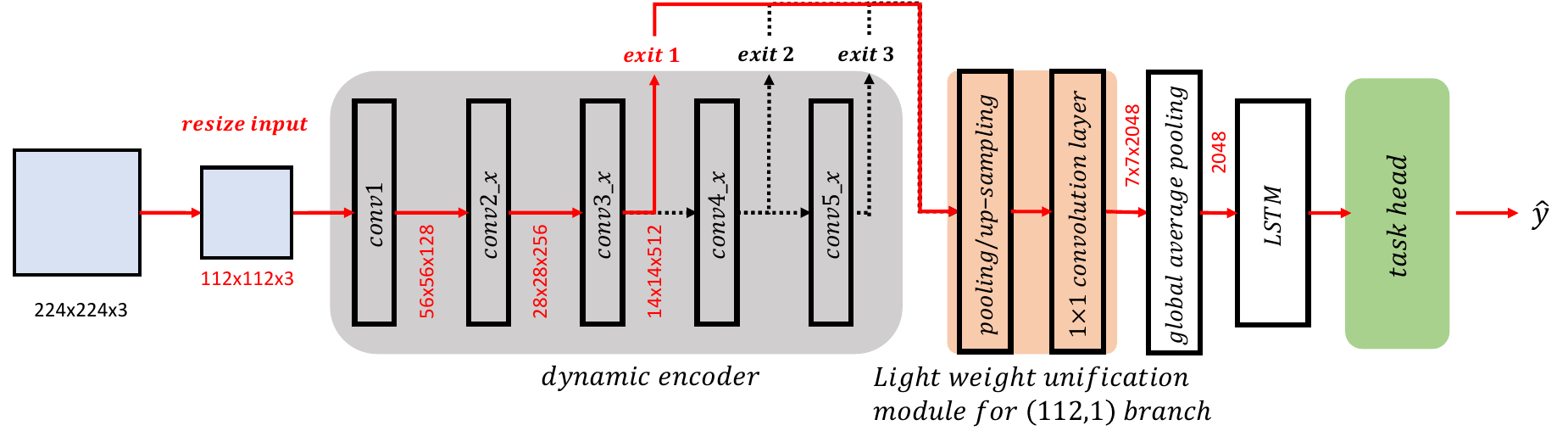}
\caption*{(a) The architecture of the main network for online action recognition}
\end{minipage}%
\\
\centering
\begin{minipage}[t]{0.8\linewidth}
\raggedleft
\includegraphics[width=1\linewidth]{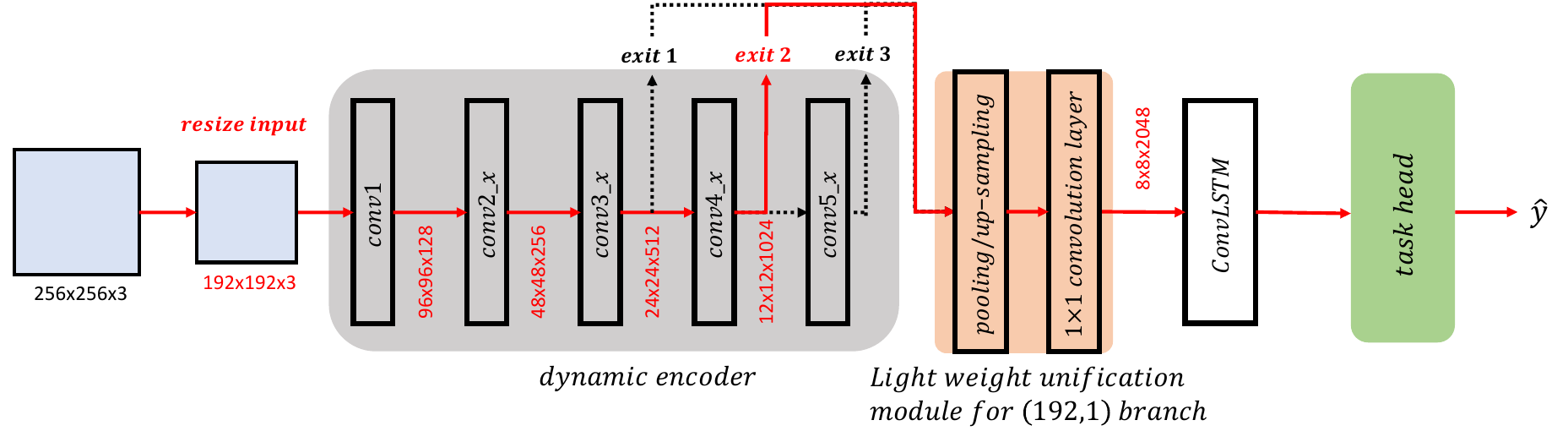}
\caption*{(b) The architecture of the main network for online pose estimation}
\end{minipage}%
\caption{The architecture of our dynamic main network. 
(a) The main network for online action recognition. The red line illustrates an example of the flow of the inputs based on the agent's policy, where the input resolution is resized to $112 \time 112$ and the $1^{st}$ exit port is activated.
(b) The main network for online pose estimation. The red line illustrates an example of the flow of the inputs based on the agent's policy, where the input is resized to $192 \time 192$ and the $2^{nd}$ exit port is activated.
}
\label{Fig: model arch}
\end{figure*}

\begin{figure}[ht]
\centering
\centering
\includegraphics[width=01\linewidth]{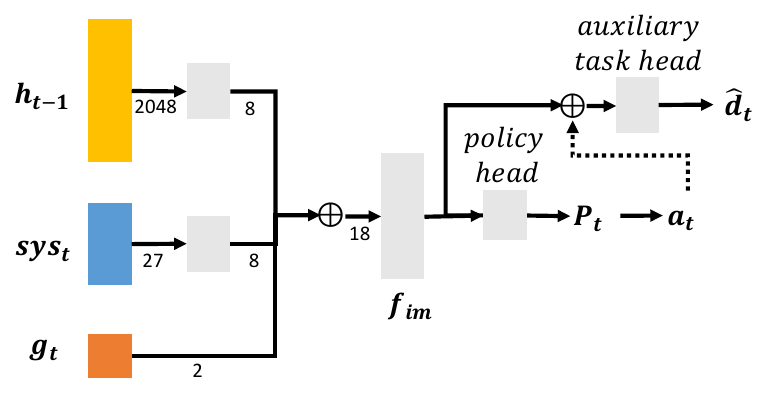}
\caption{
Illustration of our agent's architecture. The system status $sys_t$ and the hidden state $h_{t-1}$ are first separately fed into linear layers with ReLU activation. Then, the outputs of these linear layers are concatenated with the other useful information $g_t$ to generate the intermediate feature $f_{im} \in \mathbb{R}^{18}$. 
Next, this feature $f_{im}$ is fed into the policy head to generate the action probability distribution $P_t$ to select action $a_t$.
To facilitate MSA, the feature $f_{im}$ is concatenated with $a_t$, and sent to the auxiliary task head to predict the delay $\hat{d}_t$. 
Both the policy head and the auxiliary task head consists of two fully-connected layers with a ReLU activation layer in between.
}
\label{Fig: agent arch}
\end{figure}

\section{More Details on Architecture of SAN}
\label{Sec: SAN architecture}
Below, we provide more details on our SAN network.

\subsection{Our dynamic main network}
Our dynamic main network adjusts its computation complexity by dynamically changing the execution depth and the input resolution, as shown in Fig.~\ref{Fig: model arch}.
Below, we introduce more details of how our network achieves dynamic depth and resolution.

\textbf{Network for online action recognition.}
For online action recognition, we build a dynamic encoder based on ResNet50, in which three exit ports are added at the end of specific convolution blocks $\{conv3\_x, conv4\_x, conv5\_x \}$ \citeSupp{he2016deep2} to achieve dynamic depth, as illustrated in Fig.~\ref{Fig: model arch}. 
By producing predictions using only the early layers of the encoder (i.e., early-exit), we can directly reduce the computational complexity and the time consumption. 
Furthermore, because we select an encoder that is fully-convolutional (ResNet50), it is also naturally able to process inputs with varying resolutions.
Specifically, the inputs are sized to one of three resoluion candidates: $[112, 168, 224]$.

Due to the varying size of the dynamic encoder's output at different exits and using different input resolutions, we design a unification module for each resolution-depth pair in order to unify the shapes of the dynamic encoder's output features.
The unification module consists of a pooling/upsampling layer and a $1 \times 1$ convolution layer, as shown in Fig.~\ref{Fig: model arch}.
Specifically, we first utilize a pooling/upsampling layer to resize the resolution of the encoder output to $7\times 7$ and then expand the channel number of the resized output to $2048$ to obtain the final unified feature $2048 \times 7 \times 7$ via a $1 \times 1$ convolutional layer.
We remark that the unification module not only resizes the shape of the encoder's output but also helps to map features from various resolution-depth pairs to a common space.
Lastly, we perform global average pooling on the unified feature to obtain the 2048-dimensional vector and feed this vector into an LSTM, whose output is fed into the task head to predict the class of the action. 
Here, the task head is a single fully connected layer.

\textbf{Network for online pose estimation.}
For online pose estimation, the structure of the dynamic encoder is similar to that of action recognition, although the resolution candidates are instead: $[128, 192, 256]$. 
Also, differently, we utilize the unification module to resize the output of the encoder to $8 \times 8 \times 2048$ and feed the unified feature into a ConvLSTM layer \citeSupp{shi2015convolutional2} for processing the dense spatial-temporal information. Moreover, our pose estimation task head, which receives the output of ConvLSTM and outputs heatmaps for pose estimation, consists of three deconvolutional layers.
The structure of the deconvolutional layers is the same as \citeSupp{xiao2018simple2}. 

\subsection{Our agent module}
For our agent module, as shown in Fig. \ref{Fig: agent arch}, we first utilize two fully connected layers to process the system information $sys_t$  and the hidden state $h_{t-1}$ in parallel. Next, we combine the outputs of these layers with additional useful information $g_t$ (i.e., the previous
step's action $a_{t-1}$ and delay $d_{t-1}$) to build the intermediate feature $f_{im}$, which is then fed into two linear layers with a ReLU activation layer in between to generate the action distribution $P_t$ to decide the action $a_t$. 
After that, in order to facilitate MSA, the intermediate feature $f_{im}$ is combined with $a_t$ and sent to the auxiliary head to predict the delay $\hat{d}_t$.

Here, we utilize the nvitop \citeSupp{nvitop2} and psutil \citeSupp{psutil2} tools to capture the list of system status information to build the system status $sys_t \in \mathbb{R}^{27}$. We display the list below.
\begin{itemize}
\item \textbf{GPU status:} GPU's current usage rate, GPU's occupied memory, GPU's available memory, GPU's fan speed, GPU's power, GPU's temperature, and the number of running processes on GPU.
\item \textbf{CPU status}: the number of CPU threads, the CPU's current load, the CPU's average load in the last 1 minute, 5 minutes, and 15 minutes, and the average temperature of the CPU.
\item \textbf{Virtual Memory status}: the percentage of used virtual memory, the percentage of the virtual memory not being used at all (zeroed) that is readily available, the active virtual memory, and the inactive virtual memory.
\item \textbf{Swap Memory status}: the total swap memory, the used swap memory, the free swap memory, the percentage of swap memory's usage, the number of bytes the system has swapped in from disk, and the number of bytes the system has swapped out from the disk.
\item \textbf{I/O status:} the number of reads, the number of writes, the number of bytes read, and the number of bytes written.
\end{itemize}

Next, we collect two pieces of information that are generated in the previous step: the delay $d_{t-1}$ and the selected action $a_{t-1}$ to build the extra information vector $g_t \in \mathbb{R}^2$. 
Moreover, for action recognition, we directly use the output of the LSTM in the previous step to build $h_{t-1} \in \mathbb{R}^{2048}$.
For pose estimation, we pass the ConvLSTM's output through a global average pooling layer to generate $h_{t-1}$.

\section{More Discussion on Related Work}
\label{Sec: related work}

\textbf{Dynamic Networks} \citeSupp{nie2019dynamic2,meng2020arnet2,fan2020adaptive2,li2022dynamic,foo2022era,wu2018blockdrop,Wang_2018_ECCV,yang2019condconv,li20212d,tang2021manifold} generally adapt their structure or parameters according to the input, i.e., they are input-aware.
Dynamic networks are mainly adopted with the aim of computational efficiency \citeSupp{nie2019dynamic2,meng2020arnet2,fan2020adaptive2,wu2018blockdrop,Wang_2018_ECCV,li20212d,wu2020intermittent,wu2020dynamic}, although they can also be used to improve accuracy \citeSupp{li2022dynamic,foo2022era}.
Existing dynamic networks can improve efficiency by dynamically selecting a cheaper subnetwork (e.g., skipping layers \citeSupp{Wang_2018_ECCV,wu2018blockdrop}, reducing channels \citeSupp{tang2021manifold}) when it is sufficient to provide a good result for the given input, thus reducing the number of redundant computations. 
However, existing input-aware dynamic networks have the \textit{same policy under all different system conditions and platforms}.
Therefore, since system status conditions can fluctuate, these input-aware dynamic networks can make suboptimal decisions in real environments.
Compared to existing dynamic networks, our contribution lies in 2 aspects: 
(1) Our SAN is the first to be both input-aware and \textit{system-status-aware}, which considers the system status conditions when making dynamic decisions.
(2) We are also the first to consider the challenging \textit{cross-platform adaptation for dynamic networks}.
In order to conveniently deploy our SAN for different platforms, we propose a novel MSA algorithm to facilitate self-supervised adaptation to a target deployment device, without the need for the original labeled data.

{\small
\bibliographystyleSupp{ieee_fullname}
\bibliographySupp{egbib_supp}
}

\end{document}